\documentclass[sigconf]{acmart}

\usepackage{booktabs}
\usepackage[caption=false,font=normalsize,labelfont=sf,textfont=sf]{subfig}
\usepackage{hyperref}
\usepackage{array}
\usepackage{threeparttable}
\usepackage{enumitem}    
\usepackage[greek,english]{babel}
\usepackage{ifthen}
\usepackage{xspace}
\usepackage{fancybox}
\usepackage{marginnote}
\usepackage{tcolorbox}
\usepackage{multirow}
\usepackage{mathtools}
\usepackage{algpseudocode, algorithm, algorithmicx}
\usepackage{color,soul}
\usepackage{graphicx}
\usepackage{amsmath} %, amssymb}
\usepackage{tikz}
\usepackage{cleveref}
\usepackage{hhline}
\usepackage{balance}

\usepackage{listings}
\usepackage{parcolumns}
\usepackage{cleveref}

\definecolor{light-gray}{gray}{0.95}
\newcommand{\code}[1]{\colorbox{light-gray}{\texttt{#1}}}

\definecolor{circled-color}{gray}{0.15}

\def\BibTeX{{\rm B\kern-.05em{\sc i\kern-.025em b}\kern-.08em
    T\kern-.1667em\lower.7ex\hbox{E}\kern-.125emX}}

\newboolean{showcomments}
\setboolean{showcomments}{true}
\ifthenelse{\boolean{showcomments}}
{
    \definecolor{myyellow}{RGB}{255, 228, 26}
    \definecolor{myblue}{RGB}{50, 50, 220}
    \definecolor{myred}{RGB}{250, 10, 10}
    \newcommand{\nb}[2]{
        {\sf
            \fcolorbox{myyellow}{orange}{\scriptsize\textbf{#1}}%
            $\blacktriangleright$%
            {\color{myred}\fontsize{7pt}{8pt}\selectfont\textbf{[[#2]]}}%
        }%
    }
}
{
    \newcommand{\nb}[2]{}
}

\usepackage{color}

\definecolor{editorGray}{rgb}{0.95, 0.95, 0.95}
\definecolor{editorOcher}{rgb}{1, 0.5, 0} % #FF7F00 -> rgb(239, 169, 0)
\definecolor{editorGreen}{rgb}{0, 0.5, 0} % #007C00 -> rgb(0, 124, 0)

\lstdefinelanguage{JavaScript}{
  morekeywords={typeof, new, true, false, catch, function, return, null, catch, switch, var, if, in, while, do, else, case, break},
  morecomment=[s]{/*}{*/},
  morecomment=[l]//,
  morestring=[b]",
  morestring=[b]'
}
\lstdefinelanguage{HTML5}{
        language=html,
        sensitive=true, 
        alsoletter={<>=-},
        otherkeywords={
        <html>, <head>, <title>, </title>, <meta, />, </head>, <body>,
        <canvas, \/canvas>, <script>, </script>, </body>, </html>, <!, html>, <style>, </style>, ><
        },  
        ndkeywords={
        =,
        charset=, id=, width=, height=,
        border:, transform:, -moz-transform:, transition-duration:, transition-property:, transition-timing-function:
        },  
        morecomment=[s]{<!--}{-->},
        tag=[s]
}

\lstdefinelanguage{CSS}{
  morekeywords={background,color,display,justify,content,font,weight,border,size,padding},
  morestring=[s]{:}{;},
  sensitive,
  morecomment=[s]{/*}{*/}
}

\DeclareMathOperator*{\argmax}{argmax}

\newcommand{\header}[1]{\par\medskip\noindent\textbf{#1.}}

\newcommand{\toolname}{\textsc{Cortex}\xspace}

\newcommand{\VizObjs}{Visual Objects}

\newcommand{\vizobj}{visual object}
\newcommand{\vizobjs}{visual objects}

\begin{document}

\title{Page Segmentation using Visual Adjacency Analysis}

\author{Mohammad Bajammal}
\affiliation{
\institution{University of British Columbia}
\streetaddress{\ }
\city{Vancouver, BC}
\country{Canada}
\postcode{\ }
}

\author{Ali Mesbah}
\affiliation{
	\institution{University of British Columbia}
	\streetaddress{\ }
	\city{Vancouver, BC}
	\country{Canada}
	\postcode{\ }
}

\begin{abstract}
Page segmentation is a web page analysis process that divides a page  
into cohesive segments, such as sidebars, headers, and footers. 
Current page segmentation approaches use either the DOM,
textual content, or rendering style information of the page.
However, these approaches have 
a number of drawbacks,
such as a large number of parameters and rigid assumptions about the page,
which negatively impact their segmentation accuracy.
We propose a novel page segmentation approach 
based on visual analysis of localized adjacency regions.
It combines DOM attributes and visual analysis to build features of a given page 
and guide an unsupervised clustering.
We evaluate our approach,
implemented in a tool called \toolname,
on 35 real-world web pages,
and examine the effectiveness and efficiency of segmentation.
The results show that, compared with state-of-the-art, \toolname achieves
an average of 156\% increase in precision and 249\% improvement in F-measure.
\end{abstract}
    

\maketitle

\section{Introduction}
Web page segmentation is the analysis process 
of dividing a web page into a coherent set of elements.
Examples of segments include sidebars,
headers, footers, to name a few. 
The basis of segmentation is that
the contents of a segment are perceived by the user
as perceptually similar. 
% Web page segmentation has been used 
% in a wide variety of testing and analysis problems, such as
% cross-browser testing~\cite{saar2016browserbite,huse2008using},
% mobile layout bugs testing and repair~\cite{mahajan2018automated,mahajan2018automated_intl},
% security testing~\cite{geng2015combating},
% and optimizing and directing crawlers~\cite{uzun2014effective,bharati2013higwget},
% to name a few.
% All these examples rely on page segmentation
% to perform their tasks.
% Segmentation provides a number of benefits in these
% scenarios, such as abstraction,
Segmentation provides a number of benefits,
such as page abstraction~\cite{uzun2014effective,bharati2013higwget}, 
localization of bugs and repairs~\cite{mahajan2018automated,mahajan2018automated_intl},
and page difference measurement~\cite{saar2016browserbite,huse2008using}.
% Accordingly,
% a reliable and accurate segmentation can have an impact on 
% many areas of software testing and analysis.

However, existing segmentation approaches 
have a number of drawbacks.
Document Object Model (DOM)-based techniques
are one way to perform segmentation
~\cite{rajkumar2012dynamic,vineel2009web,kang2010repetition}.
In this case, data is extracted from the DOM
and then various forms of analysis are performed to identify
patterns in the DOM.
While information gained from the DOM can be useful,
these approaches, however, have one key drawback.
The analysis performed is not necessarily related
to what the user is perceiving on screen,
and therefore the number of false positives
or false negatives can be high.
An alternative approach uses text-based information~\cite{kohlschutter2008densitometric, kolcz2007site}.
In this case, only textual nodes in the DOM are extracted
as a flat (i.e., non-tree) set of strings.
Various forms of analysis,
typically linguistic in nature,
are then applied to the textual data to identify
suitable segments.
While text and linguistic information is certainly
an aspect that the user can observe,
these approaches, by definition,
do not consider other important aspects of the page,
such as style, page layout and images.
Finally, another approach uses visual DOM properties 
to perform segmentation.
This is exemplified by the VIPS algorithm~\cite{cai2003vips},
a popular state-of-the-art segmentation technique~\cite{sleiman2013survey,campus2011web}.
Although VIPS stands for Vision-based Page Segmentation,
the technique only uses visual \emph{attributes} from the DOM 
(e.g., background color) in its analysis.
It does not perform a visual analysis of the page itself from a computer vision perspective,
such as analyzing the overall visual layout.
It also makes rigid assumptions about the design of a web page.
For instance, it assumes \code{<hr>} tags always behave as horizontal rules,
and therefore their approach segments the page when it sees that tag.
Such hard coded rules result in a fragile approach with reduced accuracy,
since developers often use tags in various non-standard ways
and combine them with various styling rules.
VIPS also requires a number of thresholds and parameters
that need to be provided by the user,
thereby increasing manual effort and reducing accuracy due to sub-optimal parameter tuning.

In this paper,
we propose a novel page segmentation approach, called \toolname, 
that combines DOM attributes and visual analysis to build features
and to provide a metric that guides clustering.
The segmentation process begins by an abstraction process that
filters and normalizes DOM nodes into abstract visual objects.
Subsequently, 
layout and formatting features are extracted from the objects. 
Finally, we build a visual adjacency neighborhood of the objects 
and use it to guide an unsupervised machine learning
clustering to construct the final segments.
Furthermore, \toolname is parameter-free,
requiring no thresholds for its operation and therefore
reduces the manual effort required and
makes the accuracy of the approach
independent of manual parameter tuning. 

We evaluate \toolname's segmentation effectiveness and efficiency
on 35 real-world web pages.
The evaluation compares \toolname with the 
state-of-the-art VIPS segmentation algorithm.
Overall, our approach is able to achieve 
an average of 156\% improvement in precision
and 249\% improvement
in F-measure, relative to the state-of-the-art.

This paper makes the following contributions:
\begin{itemize}
    \item A novel, parameter-free, segmentation technique that combines both the DOM and 
            visual analysis for building features and guiding an unsupervised clustering.
    \item An implementation of our approach,
          available in a tool called \toolname.
    \item A quantitative evaluation of \toolname in terms of
          segmentation effectiveness and efficiency.
\end{itemize}

% !TEX root =  paper.tex

\section{Background and Motivating Example}
\label{sec:background}
\begin{figure}
    \centering
    {%
    \setlength{\fboxsep}{0pt}%
    \setlength{\fboxrule}{1pt}%
    \fbox{\includegraphics[trim=0 0 0 0,clip,scale=0.31]{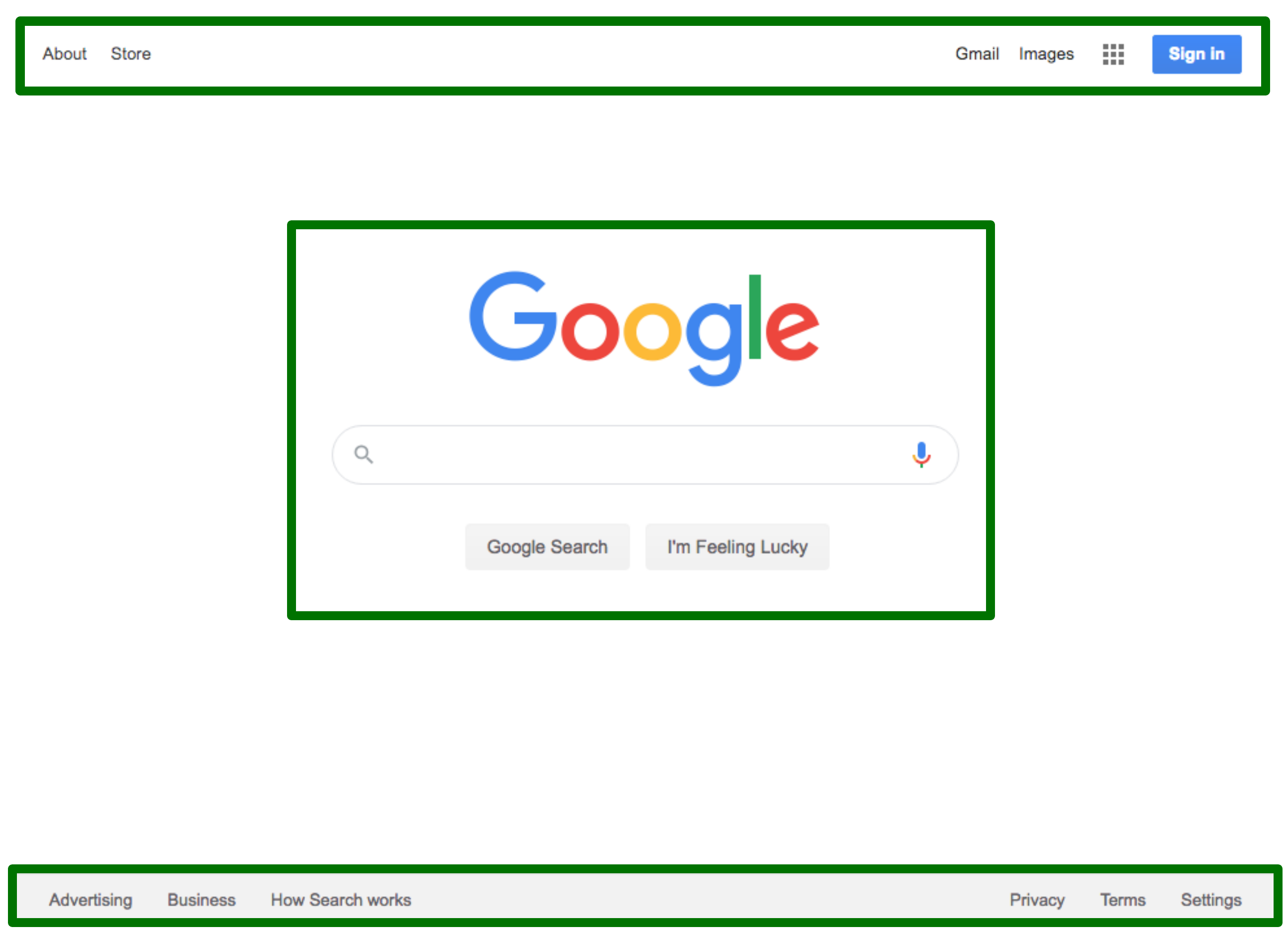}}
    }%
    \caption{An example of web page segmentation.
    Green boxes indicate detected page segments.}
    \label{fig:motivating-example}
\end{figure}

\Cref{fig:motivating-example} shows an example
of a web page with overlaid segments (marked as green boxes).
As can be seen from the figure,
the segments divide the page into a set 
of coherent groups.
Coherency in this context indicates a perceptual grouping of related elements,
where a user is able to intuitively recognize that
a page is composed of a group of segments.
For instance, in \Cref{fig:motivating-example},
a user can intuitively divide the page into 
a set of segments, such as a top/header segment, a main content segment,
and a footer segment. 

Web page segmentation is used in various areas of software engineering.
Saar et al.~\cite{saar2016browserbite}
use segmentation to test
cross-browser compatibility of web pages.
Their approach is based on
loading the same web page in two different browsers,
followed by segmenting the rendered pages in the two different browsers,
and finally comparing the pairs of segments
to ensure both pages have been rendered
in the same fashion in both browsers. 
A similar technique is used by Huse et al.~\cite{huse2008using}.
Mahajan et al.~\cite{mahajan2018automated}
propose an approach to automatically test and repair mobile layout bugs.
They first perform a segmentation of the  page to localize bugs.
Each segment is then passed to an oracle that reports a list of layout bugs.
Finally, the segment's CSS code is then patched based on a list of database patches.
A similar analysis is used for testing and repairing
web page internationalization bugs~\cite{mahajan2018automated_intl}.
Page segmentation has also been used in security testing.
Geng et al.~\cite{geng2015combating} propose a segmentation-based 
approach to detect phishing security attacks.
Their technique extracts segments from a page, and then uses
the segments to extract features, build a fingerprint of the page, 
and detect whether a page under test is phishing.

The segments shown in \Cref{fig:motivating-example} can be generated using
a number of techniques, as described in the following subsections.

\subsection{DOM-based Page Segmentation}
One approach is to use information based on the Document Object Model (DOM)~
\cite{rajkumar2012dynamic,vineel2009web,kang2010repetition}.
This approach utilizes the DOM tags, attributes, or subtrees for its analysis,
after which a set of thresholds are applied to generate a subset of DOM elements
representing the final extracted segments.
For instance, Rajkumar et al.~\cite{rajkumar2012dynamic} propose an algorithm based
on detecting tag name repetitions in the DOM.
It represents each DOM element as a string of tag names in a similar fashion
to XPaths. It then detects repeating substrings.
These repetitions (of a certain length and a certain occurrence threshold)
would then be considered web page segments.
Vineel et al.~\cite{vineel2009web} 
analyzes the DOM by first thresholding elements containing more than a certain number
of child node characters, followed by thresholding elements with more repetitive
children tag names. The rationale being that elements containing more uniform tag name
repetitions are more likely to represent a page structure.
The set of thresholded elements are then taken as the page segments.

DOM approaches, however, focus exclusively on the tag tree structure
and therefore not directly related
to what the user is actually perceiving on screen.
That is, the analysis is conducted on the tree structure by checking a set of
rules or relationships between various nodes, parents, and children.
This tree structure and the various rules and relationships between nodes
are not directly related to the final visual rendering perceived by the user.

\subsection{Text-based Page Segmentation}
A number of alternative approaches were proposed to explore complementary
ways by which the generated segments can be made more accurate and meaningful.
One such alternative approach relies on the use of text-based algorithms~
\cite{kohlschutter2008densitometric, kolcz2007site}.
This form of segmentation analyzes the textual content of
the page as opposed to the DOM tree structure.
For instance, Kohlsch{\"u}tter et al.~\cite{kohlschutter2008densitometric} divide the page into a set of text blocks.
Each block is a continuous piece of text, potentially spanning multiple tags.
The approach then computes text density, a common measure from the field of 
quantitative linguistics. It is computed by dividing the number of text tokens
by the number of lines. This is done for each text block.
Whenever two consecutive blocks have a text density difference below a certain threshold,
the blocks are merged together. This process is repeated and the resultant
blocks are taken as the page segments.
Kolcz et al.~\cite{kolcz2007site} propose an approach that first selects the text
child nodes in a predefined set of tags (e.g., \code{<table>}, \code{<div>}, \code{<blockquote>}). This excludes
certain tags that are not likely to contain significant textual information (e.g., \code{<b>}, \code{<u>}).
Next, selection is reduced to a set of text nodes that have at least
40 characters and three different types of textual tokens (e.g., nouns, verbs).
The resulting set of text blocks are taken as the final page segments.

While text-based approaches do consider an aspect of the page that is more
perceptible by the end user (i.e., the text and its characteristics),
they ignore many aspects of the page such as structure, styles, layout, and images.

\subsection{Visual Page Segmentation}
Another approach considers visual attributes of the page.
Cai et al.~\cite{cai2003vips} propose
the VIPS (\textbf{Vi}sion-based \textbf{P}age \textbf{S}egmentation)
algorithm, a quite popular state-of-the-art page segmentation tool~\cite{sleiman2013survey, campus2011web}.
The approach begins at the root DOM node.
It then iteratively splits the page to smaller segments.
Splitting is based on many hard-coded rule sets.
For example, one rule is that if a DOM node has an \code{<hr>} child, 
which represents a horizontal line,
then divide it in two (at the \code{<hr>} child) . 
The approach contains many similar hard-coded rules, 
but this makes it less robust due to assuming that developers always 
use certain tags in the same pre-defined way, which is not always true.
The approach also requires a number of thresholds,
such as a \emph{coherence threshold} that indicates whether a segment is coherent,
as well as thresholds on the dimensions of segments (e.g., width, height), among others.
Requiring many parameters from the user increases manual effort and 
often reduces accuracy due to sub-optimal parameter tuning and overfitting.

Note that the VIPS approach, despite its name, is actually not vision-based
in the sense that it does not perform visual analyses
from a computer vision perspective,
such as visually analyzing the overall visual structure of the page.
Rather, most of the analyses conducted in VIPS rely heavily on the DOM tree structure.
It was referred to as vision-based because, in some of its stages,
it uses DOM attributes that are visual in nature,
such as background color and element size.
If we envision a spectrum of techniques with DOM-based segmentation on one end
and visual segmentation on the other end, VIPS would be closer to a
DOM-based segmentation.

Visual techniques can also be at a disadvantage in some tasks. For instance, visually identifying
text blocks (i.e., via OCR - optical character recognition) can sometimes be 
inaccurate and remains an active area of research in computer vision.
On the other hand, the same task (i.e., text block identification) 
is more readily available and accessible from the DOM, 
and therefore DOM-based approaches would be more reliable in this case.

% !TEX root =  paper.tex
% \newpage
% \  \ 
% \newpage
\section{Proposed Approach}
\label{sec:approach}

The proposed approach performs web page segmentation 
based on visual analysis of the page.
Existing state-of-the-art techniques (e.g., VIPS~\cite{cai2003vips}) 
are heavily based on DOM information (e.g., element tree relationships) 
with a few visual attributes.
In contrast, our approach performs an extensive visual analysis that examines
the overall visual structure and layout of the page, 
and therefore aims to more faithfully capture the visual structure of the page as 
would be perceived by a human user, as opposed to heavily relying on how the 
elements are structured in the DOM. 
While the proposed approach is chiefly visual in nature,
it does combine aspects of both the DOM and visual page analysis 
in a fashion that aims to minimize the drawbacks of each approach, 
which were described in \Cref{sec:background}.
The approach is also parameter-free,
requiring no thresholds for its operation and therefore
reduces the manual effort required and
makes the accuracy of the approach
independent of manual parameter tuning. 

\begin{figure}
    \centering
    {%
    \setlength{\fboxsep}{0pt}%
    \setlength{\fboxrule}{1pt}%
    \fbox{\includegraphics[trim=0 0 0 0,clip,scale=0.6]{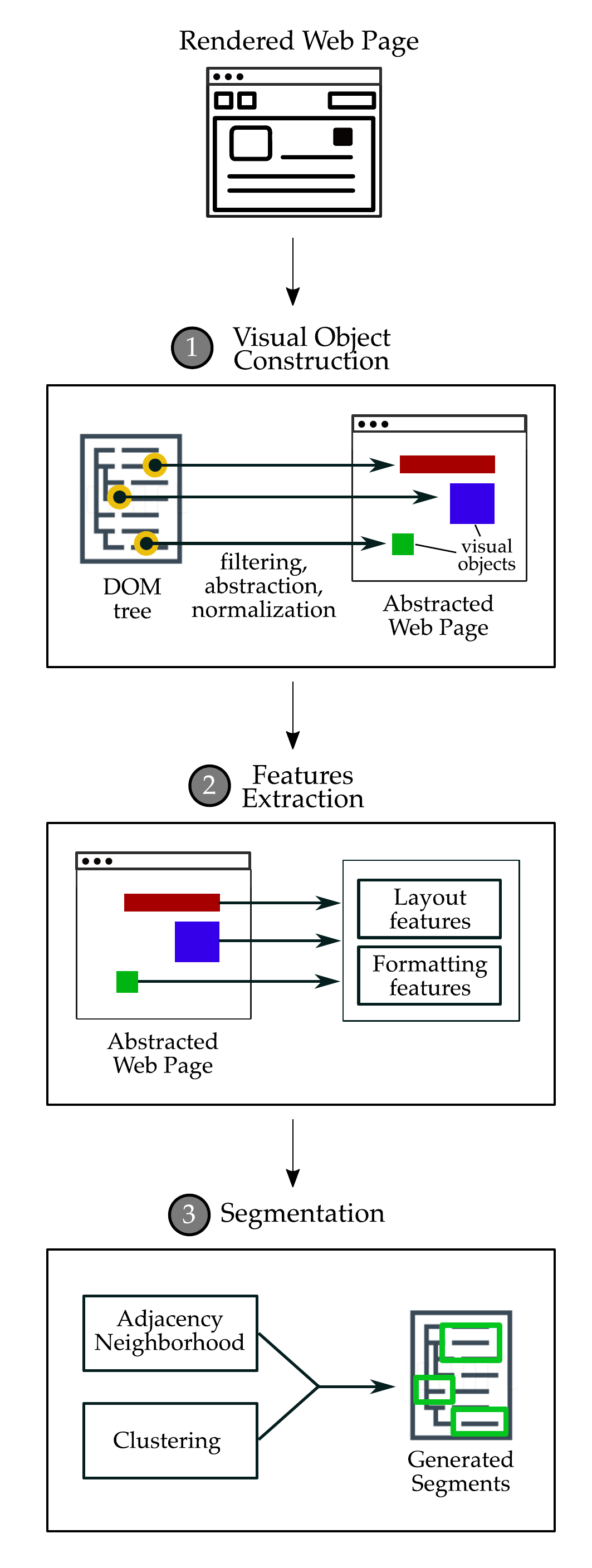}}
    }%
    \caption{Overview of the proposed approach.}
    \label{fig:approach-overview}
\end{figure}

\Cref{fig:approach-overview} shows an overview of the proposed approach.
The approach begins by retrieving the DOM 
of the rendered page.
Next, unlike techniques that are heavily based on DOM hierarchy and other DOM attributes, 
we only use a few key nodes of the DOM (as described in \Cref{subsec:abstraction}) 
and discard the rest of the tree.
The output of this process is a normalized and abstract representation of the page.
This transforms the page into a set of \vizobjs,
each of which represents a basic unit of visual information 
(e.g., a text, an image).
The approach then extracts features from these \vizobjs,
consisting of both DOM features as well as visual features.
Finally, the objects are grouped using unsupervised machine learning clustering
and the relevant DOM nodes are finally extracted as segments of the page.

In the following subsections,
we describe each step of the proposed approach
and illustrate their major components and analysis procedures.

\subsection{Visual Object Abstraction}
\label{subsec:abstraction}
In the first step of the approach,
we take as input the DOM of the page
after it is loaded and rendered in a browser.
We then perform a \emph{visual abstraction} that 
transforms the DOM into a set of \emph{\vizobjs},
which are visual abstractions of the visible subset of DOM elements.
Each {\vizobj} contains only the location and type an element.
All other information is removed. This is contrast to techniques 
that are heavily DOM-based (e.g., VIPS), which rely on DOM hierarchy 
traversal at every step of their analysis.

The rationale for this abstraction step is as follows.
First, by performing an abstraction
we aim to normalize the rendering of a page
into an abstract representation 
that signifies the salient features of the page
from a visual perspective.
The intuition behind this is that
normalization and abstraction can be helpful to 
achieve our goal of detecting segments, 
since the exact and minute page rendering details
are less relevant when aiming to divide the page as a whole
into a set of segments. 
Therefore, this visual object abstraction stage
enables obtaining a big picture overview of the page
to identify such commonalities despite minute differences.

% Furthermore,
% there is evidence that humans represent objects in an abstract way 
% before visually analyzing their grouping~\cite{humphreys2017visual, feldman2003visual}.
% That is, the details of the object (i.e., the actual picture of an image element, the meaning of a text)
% are abstracted away in favor of higher level information 
% (i.e., that it's type is a text, that it is located in a certain region).
% Our visual object abstraction behaves in the same way,
% by removing all details of the object except for high level information such as object type and location.
% % The abstraction allows the brain to easily identify similarities 
% % between various instances of an object as well as recognizing various
% % visual data as belonging to a group.

The visual object abstraction is implemented as follows.
First, we extract from the DOM a set of nodes that 
represent visual content of the page, 
and we refer to each of these as \emph{{\VizObjs}}.
We define three types of {\VizObjs}:
textual, image, and interactive.

\header{Textual Objects}
The extraction of text content is achieved by 
traversing text nodes of the DOM. More specifically:
\begin{align}
    \Theta_{t} \coloneqq \{ E \:\vert\: \nu(E) \land \tau(E) \}
\end{align}
where $\Theta_{t}$ is the set of all {\vizobjs} that represent text in the page,
$E \in DOM$ is a leaf element iterator of the rendered DOM in the browser,
$\nu(E)$ is a heuristic predicate that runs a series of checks
to detect visible elements,
and $\tau(E)$ is a predicate that examines whether
there is a text associated with $E$. 
More specifically, it returns
non-empty nodes of DOM type \code{\#TEXT},
which represent string literals. 
We note that the predicate is based on a node type, rather than
an element (i.e., tag) type.
This allows more robust abstraction because the predicate captures any text and does not
make assumptions about how developers choose to place their text.
In other words, regardless of the tag used for text data (e.g., \code{<span>, <div>}),
text would still be stored in nodes of type \code{\#TEXT}, even for custom HTML elements.
This helps in making the approach more robust by reducing assumptions about
tags and how they are used in the page.

\header{Image Objects}
Subsequently, we perform another extraction for image content.
We define this as follows:
\begin{align}
    \Theta_{m} \coloneqq \{ E \:\vert\: \nu(E) \land \mu(E) \}
\end{align}
where $\Theta_{m}$ is the set of all {\vizobjs} that represent images.
As in the previous case,
the predicate $\mu(E)$ examines whether
there is any relevant image content associated with $E$.
This has two possibilities:
nodes of \code{<img>}, \code{<svg>}, and \code{<canvas>} elements, 
and non-image nodes with a non-null background image.
We note that this predicate makes the proposed approach more robust
by eliminating assumptions about how developers choose to add images.
If images are contained in standard image tags (e.g., \code{<img>}, \code{<svg>}),
then our predicate readily captures those elements.
However, we make no assumptions that this is the only way an image can be included.
For this reason, we also capture elements of any tag type when we detect a non-null background image.

\header{Interaction Objects}
Finally, we extract the interaction elements as follows:
\begin{align}
    \Theta_{i} \coloneqq \{ E \:\vert\: \nu(E) \land \eta(E) \}
\end{align}
where $\Theta_{i}$ is the set of all {\vizobjs} that represent form elements or similar interactive elements.
These are determined by the predicate $\eta(E)$, which collects
elements such as input fields and drop down menus.

We finally obtain the total set of {\vizobjs}
in the page, $\Omega$:
\begin{align}
\Omega = \left( \Theta_t \cup \Theta_m \cup \Theta_i \right)
\end{align}

We now make a number of remarks about the abstraction process.
We use a DOM approach instead of a visual approach 
for this abstraction step for the following reasons.
While visual techniques might be useful for analyzing
the visual structure of a page since 
they mimic what a human user would be seeing,
they can be at a disadvantage in some tasks.
For instance, identifying textual objects using a visual approach is based on OCR (optical character recognition),
which involves analyzing image pixels and detecting wether or not the pixels constitute a text.
OCR remains a challenging and active area of research in the computer vision community.
The same task (i.e., identifying textual objects) 
is readily available and immediately accessible from the DOM, 
and therefore DOM-based approaches would be more suitable for this task.

Furthermore, while state-of-the-art techniques (e.g., VIPS~\cite{cai2003vips})
rely heavily on the DOM tree by traversing all elements of the tree and checking for various rules and heuristics
between parents, children, and other nodes, our approach is agnostic to the DOM tree.
Our approach does not traverse the elements of the tree and does not check for relationships between any nodes.
The approach only accesses a subset of leaf nodes, and only gets basic information from those nodes, such as node type.
The approach is therefore only loosely related to leaf nodes and agnostic to the DOM tree itself.
This observation, coupled with the fact that we use visual analysis for the remaining steps of the approach,
minimizes some of the drawbacks of DOM-based approaches mentioned in \Cref{sec:background},
such as the fact that they are not directly related
to what the user is actually perceiving on the screen.

\subsection{Features Extraction}
So far, the DOM has been
abstracted and a set of {\vizobjs} were constructed.
We now proceed by defining a mechanism to utilize these {\vizobjs}
and build on them to construct the final page segments,
which are the end goal of our proposed approach.
Accordingly, in this stage we transform each {\vizobj} constructed
in the previous stage into a feature vector.
This acts as a dimensionality reduction step
in which the {\vizobjs} are further abstracted to facilitate
reasoning and analysis. This will then be used
in subsequent stages to segment the page.

% At this feature extraction stage,
% we again utilize cues from the human brain's visual cognition system.
% Studies have indicated that,
% among other factors,
% the brain relies on cues of location, color, and continuation
% (i.e., alignments) \cite{feldman2003visual,humphreys2017visual,palmer1994rethinking}.
% These cues allow the brain to perceive 
% visual stimuli as belonging to the same object.
% We hence adopt this aspect of cognitive processing and 
% build a feature vector containing these cues.
% The features will be used in a later stage to finally
% create page segments.

We now describe the details of extracting the feature vector, 
which consists of the location, dimensions, foreground color, and background color
of each \vizobj. 
First, we extract spatial data.
We capture the x and y coordinates of the CSS box model of the \vizobj.
These are not the coordinates as defined in the DOM attributes,
but rather the \emph{computed} coordinates, as rendered by the browser.
This represents the final absolute (relative to the viewport) location
of the final rendered elements, in order to more faithfully capture the 
final visual representation as seen by the user.
We also capture the computed width and height of the box model in the same fashion.
Next, we extract color information for the \vizobjs.
Two color values are captured: background and foreground colors.
These colors are obtained through a combination of computed DOM style
values as well as computer vision methods.
The definition of these values depends on the type of the \vizobj.
For all object types, 
the background color is computed through computer vision.
The value of the background color is set to the value of the color mode 
of the region surrounding the box model. 
We use computer vision because DOM colors are declarative in nature
and do not capture the actual final rendered pixels on screen.
For instance, the computed style may indicate that the background
is transparent, while the final rendered color might actually end up being 
not transparent due to interactions with other elements of the DOM. 
This results in a situation where the computed style of 
the element itself can not be used to determine the actual rendered style.
Therefore, we use computer vision as the ultimate source of truth for 
information on the final rendered image.
For text and input objects, the foreground color is obtained from
the computed DOM style as it faithfully represents the image rendering.
For image objects, the foreground color is computed as
the color mode of the region contained inside the object's box model. 

\subsection{Page Segment Generation}

\header{Adjacency Neighborhood Construction}
In order to start analyzing the extracted {\vizobjs} features
and build page segments, we define some notion of 
adjacency information.
Adjacent objects are more likely (but not necessarily) to belong
to the same segment, and therefore it would be beneficial
to obtain some form of \emph{adjacency neighborhood} for 
the \vizobjs.
Whether or not adjacent objects actually
end up belonging to the same segment depends on the rest of the features.

The adjacency neighborhood is a data structure that captures the spatial visual layout grouping 
of the objects as rendered on the page. 
We build adjacency information using the computational geometry~\cite{toth2017handbook}
techniques often used in computer vision, which perform extensive analysis 
of how objects are overlaid with respect to each other and provide
information about their neighborhood.
The adjacency neighborhood would then be used at a later stage
to guide the unsupervised clustering process.

% We perform this step to model the brain's visual cognitive processing.
% It has been shown that spatial proximity is a strong factor
% in the brain's visual cognition when performing
% a perceptual grouping of objects
% ~\cite{palmer1994rethinking, palmeri2004visual}.
% Visual stimuli that are in close proximity to one another
% are registered in the brain as bearing more potential for grouping.
% Therefore, in this stage we model this processing mechanism in the brain
% by means of a neighborhood analysis of the extracted {\vizobjs}.

We now precisely define the adjacency neighborhood 
and the process of constructing it.
We begin by populating a spatial index from the coordinates
of {\vizobjs}.
A spatial index~\cite{beckmann1990r}
is a data structure that facilitates querying spatial relationships
between the contents of the index.
We therefore use the spatial index to resolve spatial queries and
construct an adjacency neighborhood for the extracted objects.
More concretely, we define the adjacency for {\vizobjs} as follows:
\begin{equation}
\alpha(o) \coloneqq \{ n \in \eta(o) \:\vert\: \eta(o) \cap \lambda(o, n) = \varnothing \}
\end{equation}
where $\alpha(o)$ is the adjacency neighborhood of the object $o$,
$\eta(o)$ is the nearest neighbors list of objects with respect to $o$,
and $\lambda(o, n)$ is the minimum distance line joining $o$ and $n$.
The equation computes the adjacency to $o$
where there is a direct non-intersecting 
visual line of sight with a neighbor.
This is achieved whenever the intersection of $\lambda(o, n)$
and the neighborhood $\eta(o)$ is the empty set as shown in the equation.
The end result is a set of objects comprising the
adjacency neighborhood of object $o$.

\header{Contextual Features Clustering}
Up to this point, we have
transformed the page into a set of {\vizobjs},
extracted relevant features from each object,
and constructed the adjacency neighborhood.
In this stage, we combine the adjacency neighborhood and
the feature vector to perform a \emph{contextual} features clustering.
In this process, we devise a variation of unsupervised clustering
that uses adjacency neighborhood as a context.
The rationale for this is that there will likely be a reduction in 
false positives and negatives if we were to localize the clustering
process to the adjacency.
We now describe the process by which we achieve this contextualization.

First, we analyze the adjacency neighborhood to extract a number of
\emph{scaling} factors.
These scaling factors guide the clustering towards using
the adjacency neighborhood as its context.
In other words, these factors can be thought of as adaptive thresholds 
that are automatically determined based on the data in order to 
better guide the clustering process.
More concretely, we have:
\begin{align}
    \sigma_D \coloneqq &\argmax_{\delta \,\in\, \Delta} \: \delta f(\delta) \\
    \mathrm{s.t.}& \:\: \Delta \coloneqq \{ d(o, n) \:\:\forall\:\: n \in \alpha(o), o \in \Omega\} \nonumber
\end{align}
where $\sigma_D$ is the \emph{distance factor},
$\Delta$ is the set of all pair-wise Euclidean distances, $d(o, n)$,
within all adjacency neighborhoods.
$n \in \alpha(o)$ is a member of the adjacency neighborhood of object $o$,
and $f(\delta)$ is the statistical frequency of the distance $\delta \in \Delta$.
The $\sigma_D$ factor represents the spatial distance density that matches
all clusters of adjacency neighborhoods.
In other words, the equation computes a \emph{weighted} statistical mode 
of pair-wise distances (restricted to adjacency neighborhoods).
This yields a distance value that represents 
the most probable spatial threshold of adjacency neighborhoods.
This factor $\sigma_D$ will be used in subsequent steps to guide the unsupervised clustering
in a parameter-free fashion. We now proceed to computing the next factor:
\begin{align}
    \sigma_A \coloneqq &\argmax_{\epsilon \,\in\, \Upsilon} \: \frac{f(\epsilon)}{\epsilon} \\
    \mathrm{s.t.}& \:\: \Upsilon \coloneqq \{ e(o, n) \:\:\forall\:\: n \in \alpha(o), o \in \Omega\} \nonumber
\end{align}
where $\sigma_A$ is the \emph{alignment factor},
$\Upsilon$ is the set of all pair-wise minimal alignment differences, $e(o, n)$,
within all adjacency neighborhoods.
$e(o, n)$ measures the smallest alignment differences between the pairs $o$ and $n$. 
It measures all potential alignment arrangements, such as left aligned, top aligned,
or center aligned objects. 
The equation measures a weighted statistical mode of all pair-wise
alignments in the adjacency neighborhood.
It optimizes alignment values that are as small as possible with as high
statistical frequency as possible.
This represents the scale of alignment between objects within adjacency neighborhoods.
% At this point, we reemphasize that the rationale for incorporating alignment
% measurements is to attempt modeling the brain's visual cognition.
% It is well known from cognitive science~\cite{pizlo1997curve,feldman1997curvilinearity}
% that the brain takes strong cues from the alignment of visual stimuli,
% and uses these cues to detect object grouping.
% We therefore believe that adopting similar cues can potentially be helpful
% for detecting page segments.

We now describe how $\sigma_A$ and $\sigma_D$ will be used to localize the clustering.
\begin{align}
D(o_A, o_B) \coloneqq \, S_D \: S_A \: \kappa(o_A, o_B)
\end{align}
where $S_D = f(\sigma_D)$ and $S_A = f(\sigma_A)$ are piece-wise functions
that clamp all distances below $\sigma_D$ and $\sigma_A$, respectively, to unity,
while keeping all other distances intact.
$\kappa(o_A, o_B)$ measures the perceptual differences in
background and foreground colors between the $o_A$ and $o_B$ using 
the CIE76 $\Delta{E}$~\cite{klein2010industrial} metric.
We chose this metric because it performs a comparison that takes into account
human visual perception of color, and therefore using this metric enables 
our approach to more faithfully mimic how a human would perceive the color.
Finally, the distance $D(o_A, o_B)$ is clustered using a density-based
clusterer (e.g., DBSCAN~\cite{ester1996density}).
Due to the scaling factors $\sigma_D$ and $\sigma_A$ that we have computed,
the density parameter for clustering simply becomes unity.
Once the clusters are obtained,
we retrieve the list of elements in each cluster, and
obtain their xpaths.
Each set of xpaths is finally reported as a segment
of the page and returned as the final output.

\subsection{Implementation}
We implemented the approach in a tool,
which we have called \toolname in reference to the brain's
cerebral cortex that plays a key role in perception and attention.
\toolname is implemented in Java.
We use the Selenium web driver to render the page in Google Chrome
and extract DOM trees and their relevant computed properties.
OpenCV is used for computer vision operations.
We also use the Apache Commons Math library for clustering and other
mathematical and numerical functions.
To make the study replicable, we have made {\toolname}'s source code 
and evaluation subjects available online~\cite{tool-and-data}.

% !TEX root =  paper.tex

\section{Evaluation}
\label{section:evaluation}
To evaluate \toolname, we conducted quantitative
studies aiming at answering the following research questions:\\

\begin{enumerate}[label=\textbf{RQ\arabic*},leftmargin=*]
    \item How effective is \toolname in segmenting web pages?
    \item How efficient is the segmentation process in \toolname?
\end{enumerate}

In the following subsections, we discuss the details of the experiments
that we designed to answer each research question,
together with the results.

\subsection{RQ1: Segmentation Effectiveness}
\label{sec:rq-effectiveness}
For the proposed segmentation approach to be useful and reliable,
it is important to comprehensively assess its effectiveness.
The result of segmentation is a set of 
generated web page segments, $\mathbf{\Psi}$.
Accordingly, we assess the segmentation effectiveness by 
measuring the overlap of each generated segment $\psi \in \mathbf{\Psi}$ to the ground truth,
as will be described in this section.

Furthermore, to put all measurements in perspective,
we compare our results against
the VIPS segmentation technique~\cite{cai2003vips},
which is a quite popular state-of-the-art page segmentation tool~\cite{sleiman2013survey, campus2011web}.
We use the popular Java implementation provided in\footnote{\url{https://github.com/tpopela/vips_java}}.
We use all default configurations, parameters, and thresholds in that implementation.
However, we utilized a different rendering engine in order to make the comparison fair.
The original VIPS configuration uses CSSBox\footnote{\url{http://cssbox.sourceforge.net/}}
as its rendering component, instead of a more common rendering engine such as that of Google Chrome or Mozilla Firefox.
CSSBox's role is equivalent to the web page rendering engine in a web browser.
However, we noted that, out of the box, 
VIPS segmentation was poor due to CSSBox's
inability to properly render modern web pages,
since CSSBox has not been maintained fast enough to keep up with the rapid pace of web technologies,
while the larger scale engines in Google Chrome or Mozilla Firefox are more up to speed.
So we thought that, in order to make the comparison fair for VIPS 
and focused on the actual algorithm rather
than the choice of rendering engine,
we configured VIPS to use the Google Chrome web browser
(via Selenium automation\footnote{\url{https://www.seleniumhq.org/}}).
Google Chrome has, of course, a better industry-quality
rendering engine compared to CSSBox, and therefore our configuration
of VIPS to use Google Chrome instead of CSSBox
is only an improvement of VIPS' quality.
Our proposed tool, \toolname, is also configured to use Google Chrome,
thereby ensuring that any differences in segmentation effectiveness
between the two tools are due to the algorithm itself,
instead of being due to any differences in rendering.

Accordingly, we obtain two sets of web page segments:
~$\mathbf{\Psi^V}$, the set of segments generated by VIPS
and $\mathbf{\Psi^C}$, the set of segments generated by \toolname.
Then, for each segment $\psi^v \in \mathbf{\Psi^V}$ and $\psi^c \in \mathbf{\Psi^C}$,
we compute the \emph{precision}, \emph{recall},
and \emph{F-measure} with respect to ground truth data,
which we now describe.

\header{Ground Truth}
In order to compute the precision and recall, we must first obtain
ground truth data against which the evaluation will be conducted.
Since the output of the proposed segmentation approach is a set of 
web page segments, the ground truth should also contain data 
delineating web page segments.

A basic approach would be to construct ground truth data ourselves
where we manually segment a set of subjects.
However, this approach is biased
and constitutes a threat to the validity of evaluation.
Therefore, we opted for an alternative approach. 
We use a publicly available third party dataset\footnote{\url{https://github.com/rkrzr/dataset-popular}}
that contains random web pages obtained from the Yahoo web directory,
with each page manually analyzed by a group of volunteers and divided into segments. 
A number of subjects had changed locations (i.e., their URLs were dead)
and therefore did not load in the browser and had to be excluded.
There were also other pages for which 
VIPS crashed with null pointer exceptions,
and therefore had to be excluded as well to keep the comparison fair to VIPS.
The final list of ground truth subjects
is shown in \Cref{tbl:evaluation-subjects}.
The table shows the number of DOM nodes and CSS size (in kilobytes)
for the subjects, which gives a rough idea of the complexity of the page.
The table also shows the number of ground truth segments identified in each subject.

\begin{table}
	\caption{Evaluation subjects' descriptive statistics}
	\centering
	%\fontsize{8pt}{9.2pt}\selectfont
	%\setlength\tabcolsep{2px}
	\begin{threeparttable}
		\bgroup
		\begin{tabular}{r r r r}
			\toprule
			\textbf{Subject\#} & \textbf{\#DOM nodes}	& \textbf{CSS size (KB)}  &  \textbf{\#segments} \\
			\toprule
1 	 & 	 164 	 & 	 87.0 	 & 	 5 	 \\
2 	 & 	 199 	 & 	 42.0 	 & 	 8 	 \\
3 	 & 	 280 	 & 	 23.8 	 & 	 7 	 \\
4 	 & 	 284 	 & 	 181.9 	 & 	 11 	 \\
5 	 & 	 346 	 & 	 13.1 	 & 	 13 	 \\
6 	 & 	 356 	 & 	 23.0 	 & 	 10 	 \\
7 	 & 	 368 	 & 	 86.2 	 & 	 11 	 \\
8 	 & 	 456 	 & 	 55.7 	 & 	 10 	 \\
9 	 & 	 469 	 & 	 303.1 	 & 	 18 	 \\
10 	 & 	 490 	 & 	 10.8 	 & 	 20 	 \\
11 	 & 	 500 	 & 	 63.5 	 & 	 9 	 \\
12	 & 	 524 	 & 	 81.9 	 & 	 9 	 \\
13 	 & 	 557 	 & 	 18.5 	 & 	 13 	 \\
14 	 & 	 611 	 & 	 40.8 	 & 	 8 	 \\
15 	 & 	 629 	 & 	 235.8 	 & 	 19 	 \\
16 	 & 	 636 	 & 	 120.4 	 & 	 16 	 \\
17 	 & 	 665 	 & 	 89.5 	 & 	 6 	 \\
18 	 & 	 714 	 & 	 111.0 	 & 	 22 	 \\
19 	 & 	 748 	 & 	 120.3 	 & 	 18 	 \\
20 	 & 	 772 	 & 	 251.5 	 & 	 12 	 \\
21 	 & 	 893 	 & 	 166.9 	 & 	 8 	 \\
22 	 & 	 908 	 & 	 57.4 	 & 	 21 	 \\
23 	 & 	 932 	 & 	 71.1 	 & 	 21 	 \\
24 	 & 	 942 	 & 	 406.4 	 & 	 16 	 \\
25 	 & 	 955 	 & 	 832.3 	 & 	 14 	 \\
26 	 & 	 1,004 	 & 	 150.9 	 & 	 15 	 \\
27 	 & 	 1,034 	 & 	 280.1 	 & 	 18 	 \\
28 	 & 	 1,068 	 & 	 179.8 	 & 	 6 	 \\
29 	 & 	 1,086 	 & 	 197.7 	 & 	 21 	 \\
30 	 & 	 1,191 	 & 	 117.5 	 & 	 15 	 \\
31 	 & 	 1,398 	 & 	 109.0 	 & 	 16 	 \\
32 	 & 	 1,667 	 & 	 12.9 	 & 	 11 	 \\
33 	 & 	 1,862 	 & 	 96.2 	 & 	 18 	 \\
34 	 & 	 2,344 	 & 	 53.5 	 & 	 12 	 \\
35 	 & 	 2,678 	 & 	 189.7 	 & 	 5 	 \\

\bottomrule

\end{tabular}
\egroup
\end{threeparttable}
\label{tbl:evaluation-subjects}
\end{table}

\header{Measurement}
The measurement of precision and recall is performed as follows.
First, for each test subject,
we obtain the set of segments $\mathbf{\Psi^C}$ generated by \toolname
as well as the set of segments $\mathbf{\Psi^V}$ from VIPS.
We also obtain the set of ground truth segments, $\gamma \in \mathbf{\Gamma}$.
Then, in order to measure the precision and recall, we need to define 
notions of true positive, false positive, and false negative outcomes.

Our definitions are based on treating each generated segment as being composed of two regions:
a true positive region, where the generated segment overlaps with the ground truth segment,
and a false positive region, where the generated segment does not overlap with the ground truth. 
More specifically:

\begin{equation}
    \label{eqn:tp}
    TP = \sum_{\psi \in \mathbf{\Psi}} \vert \psi \cap \eta(\psi) \vert
\end{equation}
where $TP$ is the true positive area and $\eta(\psi) \in \mathbf{\Gamma}$ is the nearest neighbor of 
output segment $\psi$, but from the ground truth set $\mathbf{\Gamma}$. 
We use a nearest neighbor search in order to identify the pairs
of segments that should be compared. 
That is, for each generated segment, which ground truth segment
should be used for measuring precision and recall.
Then, after this pairing, \cref{eqn:tp} sums the areas of intersection
for all pairs. This total area becomes the true positive measure. 

As for false positives, we use the following measure:
\begin{equation}
    \label{eqn:fp}
    FP = \sum_{\psi \in \mathbf{\Psi}} \vert \psi - \eta(\psi) \vert
\end{equation}
in this case, we also begin by finding the nearest neighbor similar
to our true positive calculation in \cref{eqn:tp}.
However, instead of taking the area of intersection between each pair,
we exclude overlapping areas between the pairs, and measure the remaining area.
This measures how much of the generated segment is \emph{not} overlapping
with ground truth, thereby indicating a false positive.
We note that, in case there is no overlap at all between
the generated segment and the ground truth,
then \cref{eqn:fp} correctly measures the \emph{entire} area
of the generated segment as false positive.

The false negative computation differs slightly from the two previous measures
for true and false positives. Because a false negative is, by definition,
absent from the generated segments,
the measurement can not iterate over the output set $\mathbf{\Psi}$ as in
the previous two equations.
Instead, the iteration is performed over the ground truth set $\mathbf{\Gamma}$,
as follows: 
\begin{equation}
    \label{eqn:fn}
    FN = \sum_{\gamma \in \mathbf{\Gamma}} \vert \gamma - \mu(\gamma) \vert
\end{equation}
where $\mu(\gamma) \in \mathbf{\Psi}$ is the nearest neighbor of 
ground truth segment $\gamma$ in the set of output segments $\mathbf{\Psi}$.
This nearest neighbor function $\mu(\gamma)$ acts in an opposite manner to the
preceding nearest neighbor function $\eta(\psi)$ used in equations \ref{eqn:tp} and \ref{eqn:fp}.
$\mu(\gamma)$ pairs a ground truth segment $\gamma$ with the matching output segment $\psi$,
while $\eta(\psi)$ pairs an output segment to a ground truth segment. 
The false negative is then measured as the non-overlapping region of a ground truth segment.
We note that, in case there is no overlap at all between
the ground truth segment and the generated segment,
then the entire area of the ground truth segment is measured as false negative.

\begin{figure}
    \centering
    {%
    \setlength{\fboxsep}{0pt}%
    \setlength{\fboxrule}{1pt}%
    \fbox{\includegraphics[trim=0 0 0 0,clip,scale=0.55]{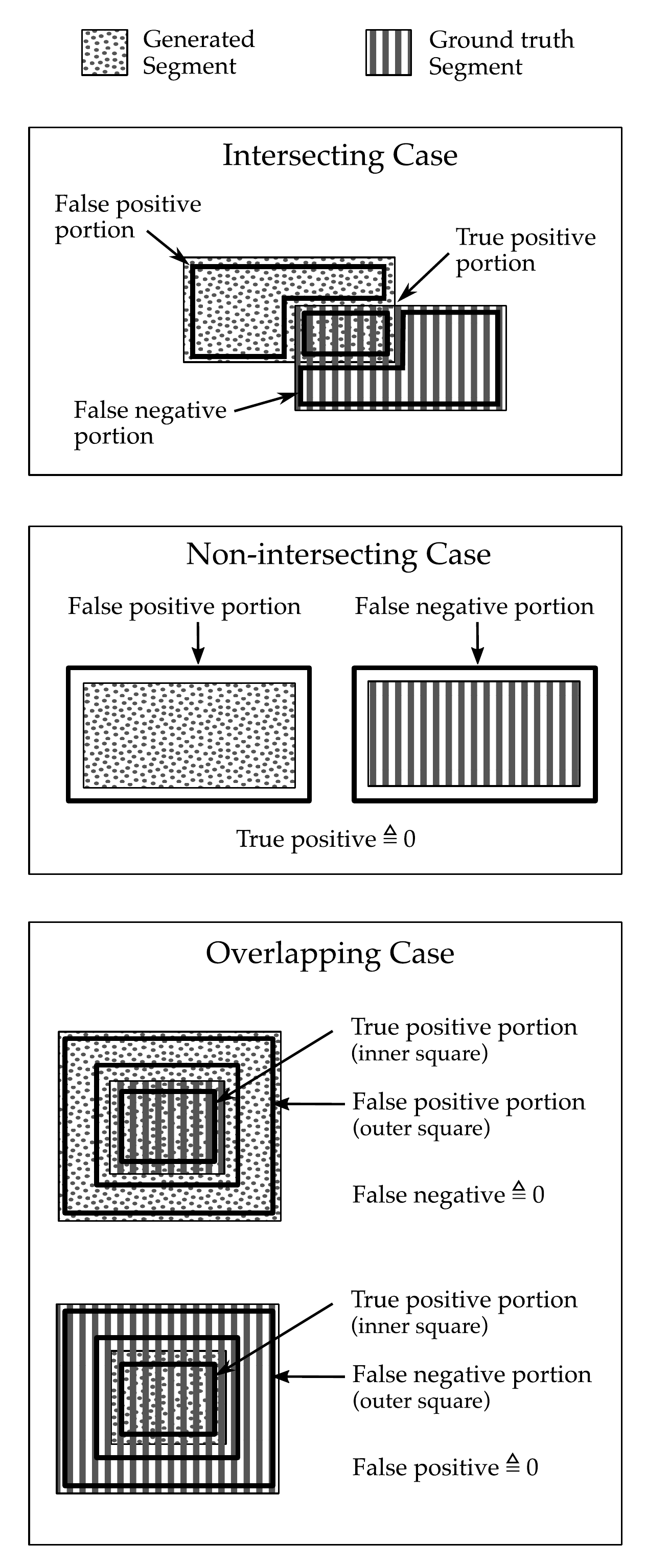}}
    }%
    \caption{Arrangement variations of
    ground truth segments and output segments
    for evaluating segmentation effectiveness.
    %fixed \mohammad{The dotted patterns appear weird dependeing on the pdf viewer. I will change the pattern later.}
    }
    \label{fig:possible-intersections}
\end{figure}

To summarize the approach we use to calculate the different measurements,
\Cref{fig:possible-intersections} shows all possible arrangements
of output segments and ground truth segments, and how the true positive, false positive,
and false negative values are defined in each case.

We make one final remark regarding these measurements.
The measurements are performed for the subject \emph{as a whole}.
That is, even though the equations iterate over each segment,
the measure is indicative of the effectiveness
of segmentation for the subject as a whole.
No averaging of any kind is performed
and therefore the final computed precision and recall
represent \emph{exact} as opposed to averaged measurements.

\subsection{RQ2: Efficiency of Segmentation}
We evaluate the efficiency of the segmentation process in order to assess
how computationally expensive is the segmentation.
While segmentation precision and recall is of prime importance,
having a highly accurate segmentation
that is computationally prohibitive makes it practically unusable for real-life web applications.
We therefore believe it is useful to complement the precision evaluation with
an examination of efficiency.

Similar to \Cref{sec:rq-effectiveness},
we also compare the efficiency of \toolname against
that of VIPS and use the same subjects.
In order to ensure a fair comparison, we make a number of remarks.
First, both VIPS and \toolname
were implemented in the same programming language (Java)
to reduce potential efficiency disparities due to programming language choice.
Second, VIPS was configured 
to use Google Chrome (via the Selenium framework),
instead of CSSBox used in the original configuration.
One reason for this choice,
in addition to rendering accuracy issues as mentioned in \Cref{sec:rq-effectiveness},
is to ensure that both tools, VIPS and \toolname,
are using the same browser engine in order to neutralize any potential
differences in efficiencies pertaining to web page rendering or DOM access.

\header{Measurement}
We measure the execution time as follows.
First, we launch browser sessions and load a test subject.
None of these steps are timed as they are outside the scope of segmentation 
and are also necessary prerequisites to any segmentation tool.
Next, we prepare and/or cast the variables in the expected classes and formats
required by each tool. None of these steps are timed either, since they
occur before the segmentation execution itself.
After all required inputs are ready, we run the segmentation and begin the timing process.
Once all segments are generated and available for use, we stop timing.

\section{Results and Discussion}
\begin{table}
	\caption{Precision, recall, and efficiency comparison}
	\centering
	%\fontsize{8pt}{9.2pt}\selectfont
	%\setlength\tabcolsep{2px}
	\begin{threeparttable}
		\bgroup
		\begin{tabular}{l r r c}
			\toprule
											& \textbf{VIPS}		& \textbf{Cortex}  & \textbf{Significance} 	\\
			\toprule
			Precision          				& 32.1\%            & \textbf{50.1}\%  & $t_{score} = $ 5.15    \\
											& 					& 				   & $p < 0.00001$ 			\\
			relative improvement          	& ---               & 156.1\%          & \ 						\\
			\midrule 
			Recall         					& 13.7\%            & \textbf{38.8}\%  & $t_{score} = $ 5.29    \\
											& 					& 				   & $p < 0.00001$ 			\\
			
			relative improvement         	& ---               & 283.2\%          & \ 						\\
			\midrule
			F-measure         				& 15.7\%            & \textbf{39.1}\%  & $t_{score} = $ 6.67    \\
											& 					& 				   & $p < 0.00001$ 			\\

			relative improvement          	& ---               & 249.0\%          & \   					\\
			\midrule
			Efficiency (seconds)  			& 57.3 s            & \textbf{0.694} s & $t_{score} = $ 7.76    \\
											& 					& 				   & $p < 0.00001$ 			\\
			relative improvement          	& ---               & 8,256.5\%        & \   					\\
			\bottomrule

		\end{tabular}
		\egroup
	\end{threeparttable}
	\label{tbl:results-summary}
\end{table}

\begin{figure}
    \centering
    {%
    \setlength{\fboxsep}{0pt}%
    \setlength{\fboxrule}{1pt}%
    \fbox{\includegraphics[trim=0 230 475 0,clip,scale=0.31]{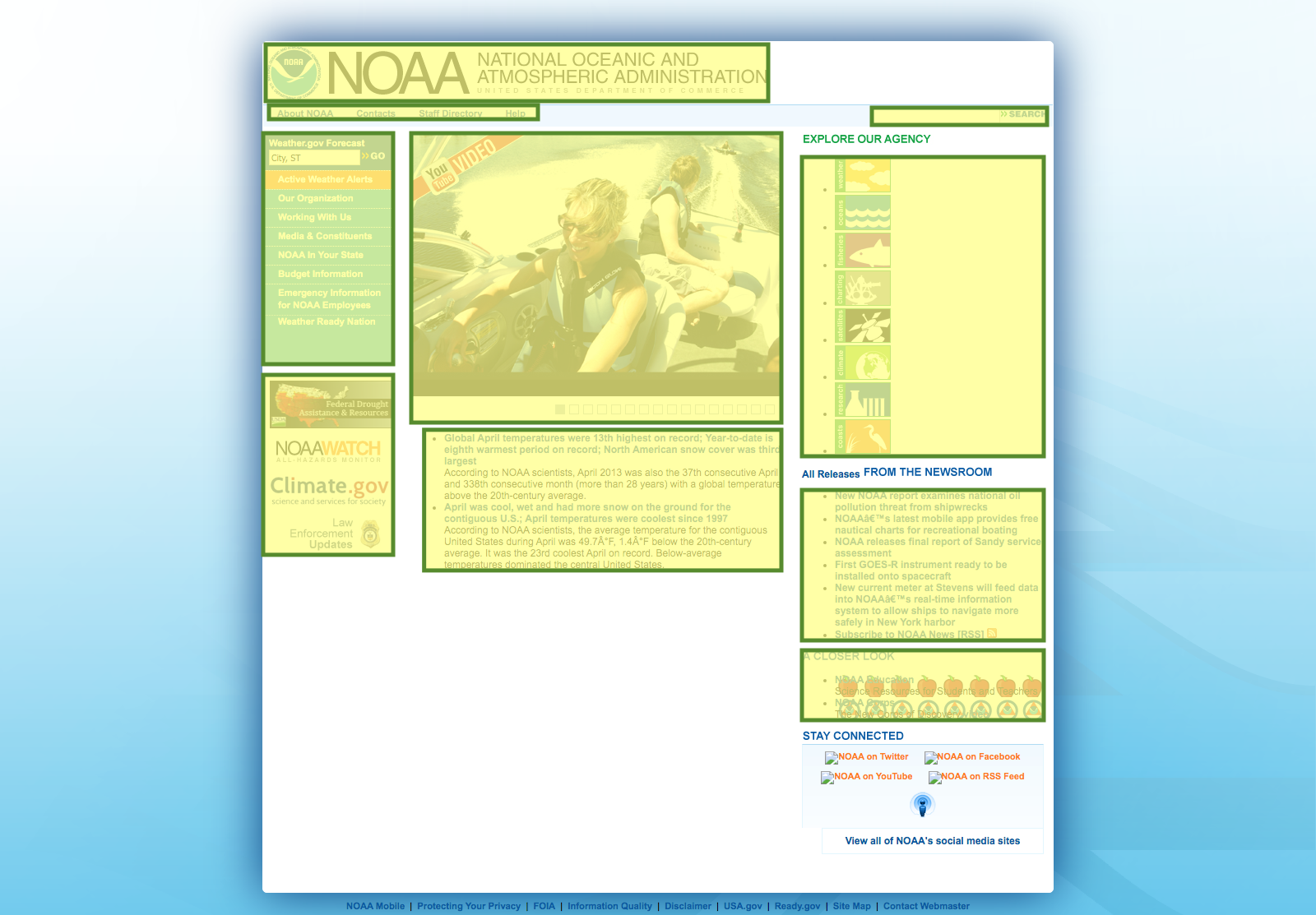}}
    }%
    \\Ground truth\\ \ \\
    {%
    \setlength{\fboxsep}{0pt}%
    \setlength{\fboxrule}{1pt}%
    \fbox{\includegraphics[trim=0 230 475 0,clip,scale=0.31]{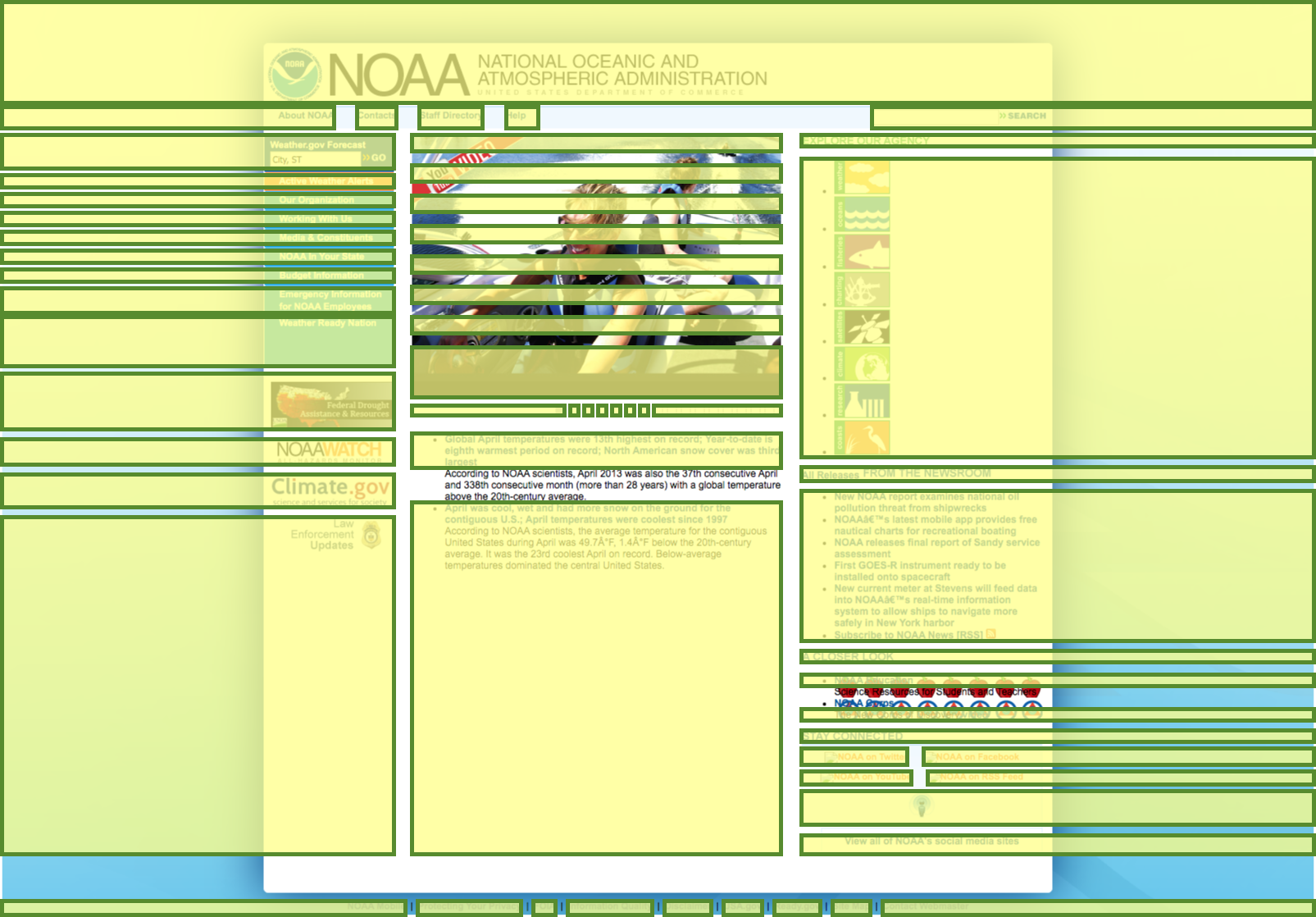}}
    }%
    \\VIPS segmentation\\ \ \\
    {%
    \setlength{\fboxsep}{0pt}%
    \setlength{\fboxrule}{1pt}%
    \fbox{\includegraphics[trim=0 230 475 0,clip,scale=0.31]{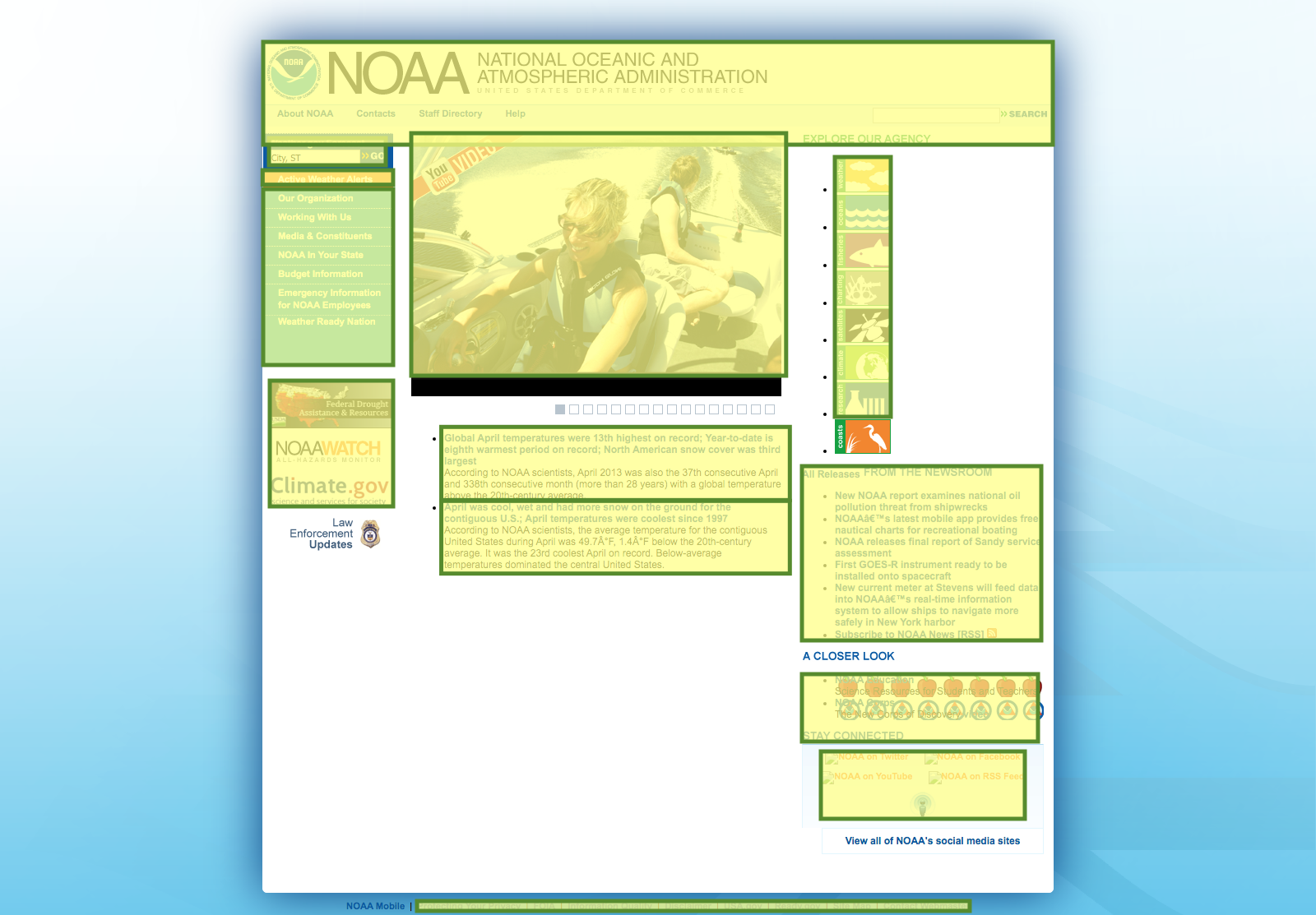}}
    }%
    \\Cortex segmentation
    \caption{Comparison of ground truth segments to the 
    segments generated by VIPS and Cortex. 
    Each yellow highlighted rectangle represents a segment. }
    \label{fig:output-comparison}
\end{figure}

\begin{figure}
    \includegraphics[trim=0 0 0 0,clip,scale=0.56]{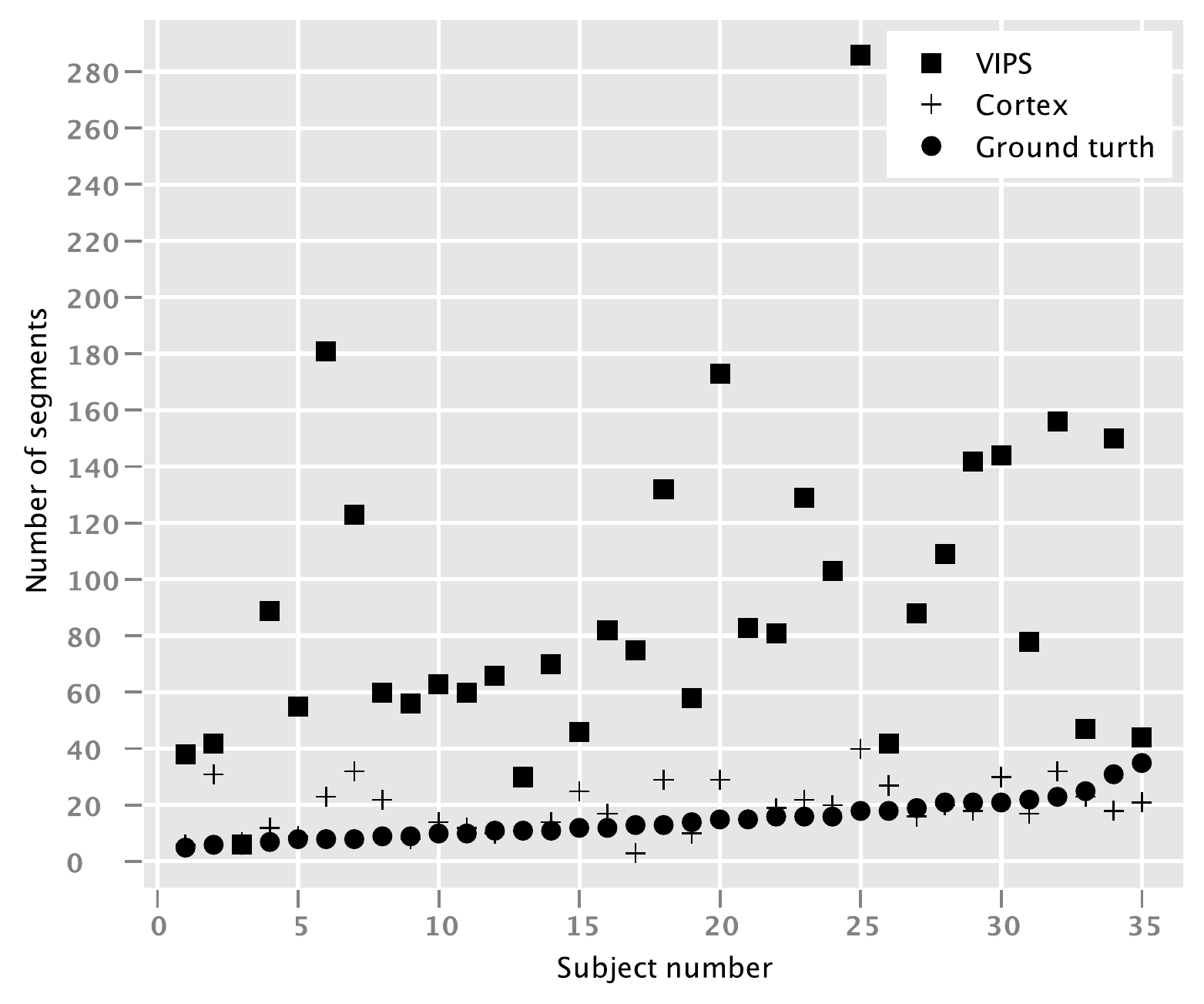}
    \caption{Number of segments generated by VIPS and Cortex, compared to ground truth.}
    \label{fig:number-of-segments}
\end{figure}

\Cref{tbl:results-summary} shows a summary of
the precision, recall, and efficiency evaluation. 
The first column shows the variable being measured,
in both its absolute value as well as relative improvement percentage
relative to baseline.
The second and third columns show the
measured evaluation average for both VIPS and Cortex,
respectively, across all subjects.
Finally, the third column tests for statistical significance
using Welch's unequal variances t-test. 
\toolname's average precision was 50.1\%
compared to 32.1\% for VIPS, showing an improvement of 156.1\%.
The result was highly statistically significant,
with a t-score of 5.15 and $p <$ 0.00001.
The recall, F-measure, and runtime
also showed an improvement of 283.2\%, 249\%, and 8,256.5\%, 
respectively, relative to VIPS.
All evaluation improvements were statistically significant with $p <$ 0.00001.

We now investigate these results in more detail
in order to understand the reasons behind
the measured differences in evaluation outcomes.
We begin by showing how the output of segmentation looks like.
\Cref{fig:output-comparison} shows the segments generated
by VIPS and \toolname for one of the evaluation subjects.
For reference, the ground truth segments for that subject
are also shown in the figure.
Each segment is represented by a rectangle
with a yellow highlight and a green border.

We make a number of observations
from the comparison of the two outputs.
First, from a big picture perspective,
it can be observed that the \toolname
output is more similar to ground truth,
compared to VIPS' output.
That is, the overall arrangement, count, and size of segments
from \toolname has better resemblance to ground truth.
This illustrates an example of the reason behind the measured
evaluation improvements for \toolname.
In order to better understand these improvements, we 
examine how VIPS is performing 
and then contrast how \toolname generates better results.
We first observe that the precision of VIPS is relatively low,
as can be seen from the typically very large sizes of the generated segments.
VIPS does not create precise segments that closely match the ground truth.
Rather, it has a tendency of oversegmentation, where almost
every element ends up being in its own segment.
\Cref{fig:number-of-segments} illustrates this behavior,
where the number of segments generated by both VIPS and \toolname
are plotted for all subjects. Note how VIPS has a tendency of generating
significantly more segments than necessary.
This can be seen, for example, in the VIPS output in \Cref{fig:output-comparison},
where there are many segments over the central image element,
and over the small buttons under it.
This might be attributed to VIPS' iterative nature
and its lack of accurate definition of visual elements,
therefore increasing the likelihood that more segments are generated.
Contrast this with the one-step approach used in \toolname.
The segments are generated in a non-iterative fashion,
and the segments originate from abstract visual objects as opposed to 
the entire DOM node set.
The final number of segments would therefore tend 
to be relatively small compared to VIPS,
which reduces the potential for false positives
and results in a relatively higher precision.

\begin{figure}
    \centering
    \includegraphics[trim=0 0 0 0,clip,scale=0.56]{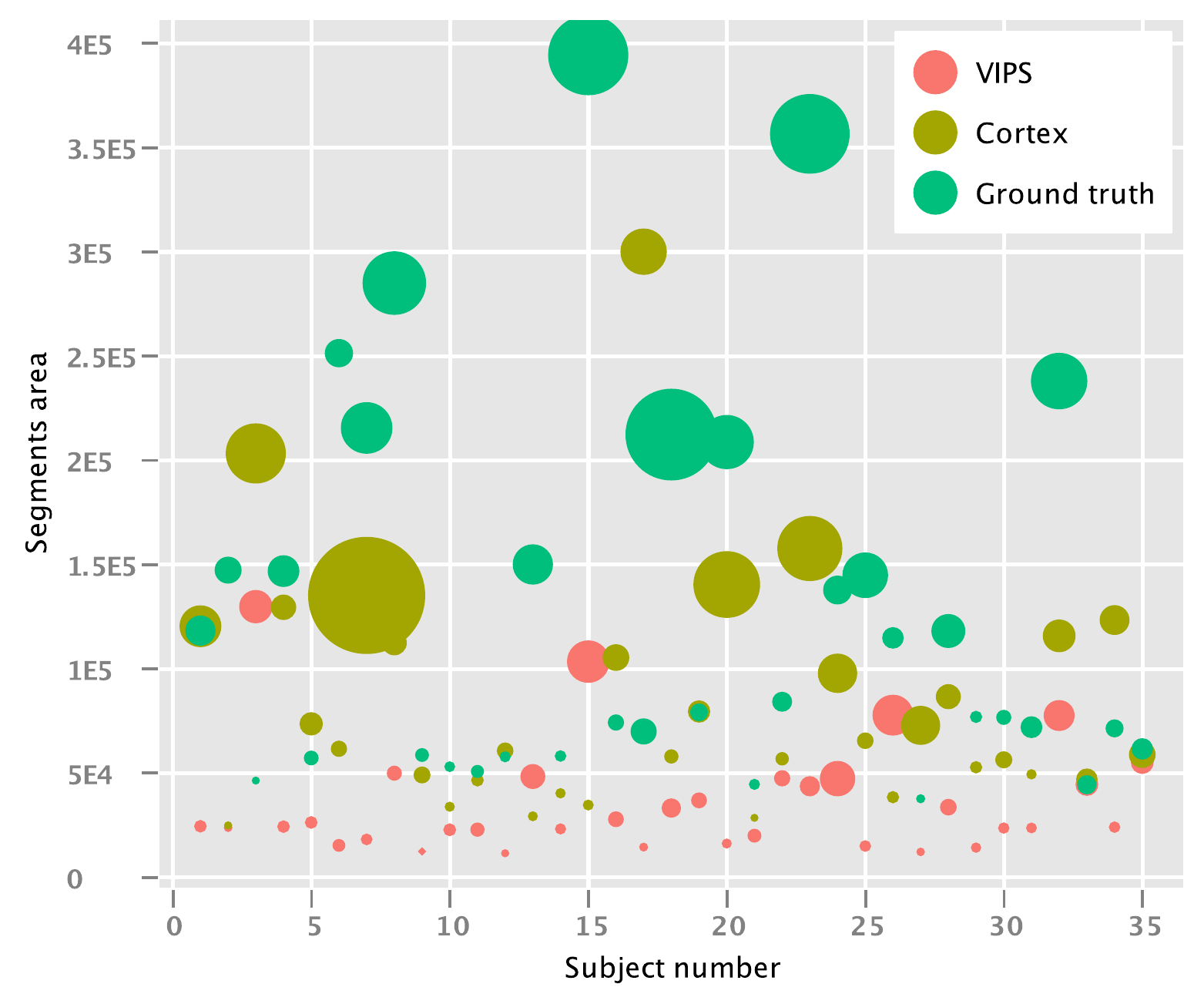}
    \caption{Generated segment areas of VIPS and Cortex compared to ground truth.
    Y value is the average segment area and bubble size is the standard deviation.}
    \label{fig:segment-area}
\end{figure}

VIPS also tends to include more elements in a segment,
resulting in larger segments and lower precision.
For instance, in \Cref{fig:output-comparison} note how the VIPS segments
of the left menu sidebar include the entire row,
resulting in a segment that includes non-visible parent elements that have greater
areas but are not visually present on the screen.
This might be attributed to VIPS' top-down approach,
where it starts from the whole page as one segment,
then iteratively divides it into smaller segments.
The final segments would therefore generally be
expected to have a tendency to be large, because the starting segment
is of maximum size (i.e., the whole page).

However, the same iterative behavior in VIPS would also equally
generate very small segments when the iteration stopping condition is not met.
This can be seen, for instance, in the VIPS output in \Cref{fig:output-comparison}
for the small button under the central image, and also the four small squares
around the top menu bar items.
\Cref{fig:segment-area} illustrates this behavior.
The figure shows a bubble plot of the segment areas for each subject.
The y-value of each bubble indicates the average segment area for that subject,
and the bubble size indicate the standard deviation of areas.
Note how \toolname generates segments that have more similar areas and distribution
as the ground truth, compared to VIPS, where the segments are mainly located
at the bottom of the plot.

Contrast this with the non-iterative approach used in \toolname.
Here we start from small leaf nodes, and form segments
by merging. As there is no stopping condition to reach,
and therefore no continuous sub-division of segments is performed,
the final segments can therefore be expected
to be on the larger end of the spectrum.
This behavior, combined with the small number of generated segments
and better overlap with ground truth,
yields a relatively higher precision for \toolname.

However, the precision and recall of \toolname,
while relatively high, is still not ideal.
We attribute this to the feature selection and the clustering process.
\toolname has a tendency, as expected, to create different segments where there is
strong and pronounced variations in color. But this sometimes leads to oversegmentation.
For instance, in \Cref{fig:output-comparison} compare the \toolname segment
over the left menu sidebar with that in ground truth.
In the ground truth,
the top most part of the sidebar has a search field,
followed by an orange menu item,
followed by the rest of the menu.
While in ground truth these are all correctly identified as one segment,
\toolname identifies each of these three parts as their own separate segments
due to very pronounced color style differences.
The same behavior can be observed in the top most segment of ground truth.
While the top most area of ground truth has a separate logo and brand segment,
and another top menu segment,
these two parts are considered one segment in the \toolname output.

\header{Threats to validity}
To mitigate potential selection bias,
we selected a random set of subjects as described in \Cref{sec:rq-effectiveness}.
Furthermore, the subjects are diverse and complex enough to be
representative of real-world web pages,
mitigating threats to the external validity of the study by making the results generalizable.
To mitigate the experimenter-bias internal threat,
we use ground truth data that is publicly available and has been collected and labeled
by an external third-party.
Furthermore, we conduct statistical significance testing
to ensure the observed outcomes are significant.

\section{Related Work}
We already discussed some of the related work on page segmentation in \autoref{sec:background}.
Here we focus on techniques that
analyze web pages from a visual perspective.
An advantage of visual analysis approaches is that
they tend to better capture what an end user would perceive.
Choudhary et al.~\cite{choudhary2012crosscheck} propose
an approach that detects cross-browser compatibility
by examining visual differences between the same page running in multiple browsers.
Bajammal et al.~\cite{canvas_icst2018} propose an approach
to analyze and test web canvas element through visual
inference of the state of the canvas and its objects,
and allowing canvas elements to be testable using common DOM testing approaches.
Stocco et al.~\cite{2018-Stocco-FSE} employ computer vision techniques
 for visual-based web test repair and migrating DOM-based tests to visual tests.
Burg et al.~\cite{burg2015explaining} present a tool that helps
developers understand the behavior of web apps.
It allows developers to specify which element they are interested in,
then tracks that element for any visual changes and the corresponding code changes.
In contrast to our work,
none of these studies aims to automatically generate segments from a web page.

\section{Conclusions}

Web page segmentation is the process of extracting
sets of cohesive elements from a web page.
It has been used in various applications,
such as cross-browser testing,
mobile layout bugs testing and repair,
security testing, and crawling optimization.
However, existing segmentation approaches,
such as DOM-based, text-based, or vision-based segmentation,
have a number of drawbacks that reduce their accuracy.
In this paper, we proposed a novel visual segmentation approach.
Unlike existing state-of-the-art techniques,
which are mainly DOM-based with a few visual attributes,
our approach performs an extensive visual analysis that examines
the overall visual structure and layout of the page, 
and therefore more faithfully captures the visual structure of the page as perceived by a human user. 
While our approach is mainly visual in nature,
it also combines aspects of both DOM-based and visual-based segmentation
in a fashion that aims to minimize the drawbacks of each segmentation technique.
Furthermore, the approach is parameter-free,
requiring no thresholds for its operation and therefore
reduces the manual effort required and
makes the accuracy of the approach
independent of manual parameter tuning. 
We implemented our approach in a tool called \toolname,
and evaluated its effectiveness and efficiency of
segmenting real-world web pages. 
It achieved an average of 156\% improvement in precision
relative to state-of-the-art, 283\% improvement in recall, 
and a 249\% improvement in F-measure.

\balance
\bibliographystyle{IEEEtran}
\bibliography{bibliography}

\end{document}